\pdfoutput=1

\documentclass[11pt]{article}

\usepackage{naacl2021}
\usepackage{graphicx}
\usepackage{tcolorbox}
\usepackage{xcolor}
\usepackage{soul}

\usepackage{times}
\usepackage{latexsym}
\usepackage{amsmath,amsfonts}

\usepackage[T1]{fontenc}

\usepackage[utf8]{inputenc}

\usepackage{microtype}

\usepackage{multirow}

\newcommand{\ourdataset}{{WikiGraphs}}
\newcommand{\secref}[1]{Section \ref{#1}}
\newcommand{\tabref}[1]{Table \ref{#1}}
\newcommand{\figref}[1]{Figure \ref{#1}}

\title{\ourdataset: A Wikipedia Text - Knowledge Graph Paired Dataset} 

\author{Luyu Wang\textsuperscript{\normalfont{*}}\and Yujia Li\textsuperscript{\normalfont{*}}\and Ozlem Aslan\and Oriol Vinyals \\
  \textsuperscript{*}Equal contribution \\
  DeepMind, London, UK \\
  \texttt{\{luyuwang,yujiali,ozlema,vinyals\}@google.com}
}

\begin{document}
\maketitle
\begin{abstract}
We present a new dataset of Wikipedia articles each paired with a knowledge graph, to facilitate the research in conditional text generation, graph generation and graph representation learning.  Existing graph-text paired datasets typically contain small graphs and short text (1 or few sentences), thus limiting the capabilities of the models that can be learned on the data.  Our new dataset \ourdataset~is collected by pairing each Wikipedia article from the established WikiText-103 benchmark \cite{merity2016pointer} with a subgraph from the Freebase knowledge graph \cite{bollacker2008freebase}.  This makes it easy to benchmark against other state-of-the-art text generative models that are capable of generating long paragraphs of coherent text.  
Both the graphs and the text data are of significantly larger scale compared to prior graph-text paired datasets. 
We present baseline graph neural network and transformer model results on our dataset for 3 tasks: graph $\rightarrow$ text generation, graph $\rightarrow$ text retrieval and text $\rightarrow$ graph retrieval. We show that better conditioning on the graph provides gains in generation and retrieval quality but there is still large room for improvement.
\footnote{The data and the code to reproduce our baseline results are available at \url{https://github.com/deepmind/deepmind-research/tree/master/wikigraphs}}
\end{abstract}

\section{Introduction}

Parallel datasets that pair data from different sources and modalities have enabled large amounts of research on cross modality learning.  Paired image-caption datasets enable models to describe visual scenes in natural language \cite{lin2014microsoft,vinyals2016show}, paired streams of speech and transcription data makes it possible to train speech recognition systems \cite{garofolo1993darpa,panayotov2015librispeech} or text-to-speech synthesis models \cite{oord2016wavenet},
and parallel corpus of text in different languages enable learned machine translation models \cite{barrault2020findings}.

\begin{figure}[t]
    \centering
    \includegraphics[width=\columnwidth]{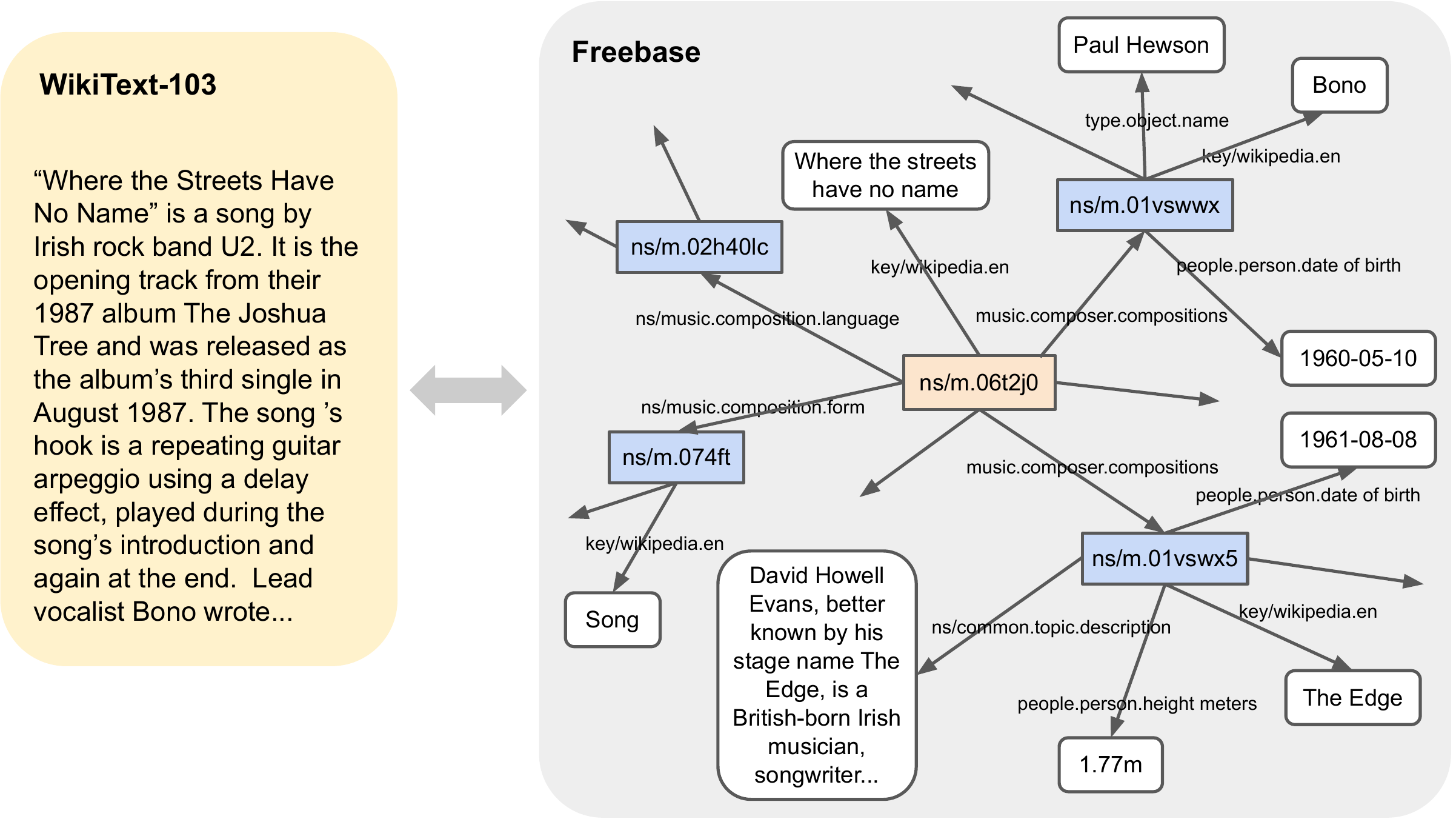}
    \caption{Illustration of a pair of Wikipedia article and the corresponding knowledge graph in our dataset.
    }
    \label{fig:cartoon-example}
\end{figure}

We present a new dataset of Wikipedia text articles each paired with a relevant knowledge graph (KG),
which enables building models that can generate long text conditioned on a graph structured overview of relevant topics,  and also models that extract or generate graphs from a text description.

There has been many prior efforts trying to build datasets for learning graph $\rightarrow$ text generation models \cite{jin-etal-2020-genwiki, gardent-etal-2017-webnlg, lebret-etal-2016-neural}.
However, existing graph-text paired datasets are mostly small scale, where the graphs tend to have 10-20 or even less nodes, and the text typically only contains one or a few sentences.  This represents a significant contrast with the state-of-the-art text generation models \cite{dai2019transformer,brown2020language}, which can already generate very fluent and long text that spans thousands of tokens over multiple paragraphs.

We attempt to bridge this gap, with the goal of advancing the state-of-the-art graph $\rightarrow$ text generation models, graph representation learning models and also text-conditioned graph generative models.  Each text document in our dataset is a full-length Wikipedia article, and we pair each of them with a KG that are significantly bigger than prior datasets of similar nature and includes much richer information.  Hand labelling text articles with KGs is expensive and not scalable \cite{jin-etal-2020-genwiki},
therefore we utilize an existing and established knowledge base, Freebase \cite{bollacker2008freebase}, and designed an automated process to extract a relevant subgraph from it for each Wikipedia article. To make the text generation results on our dataset directly comparable to the state-of-the-art, we chose the set of Wikipedia articles from the established language modeling
benchmark WikiText-103 \cite{merity2016pointer}, which contains a subset of high-quality Wikipedia articles.  This gives us a dataset of 23,522 graph-text pairs in total, covering 82.3\% of Wikitext-103 articles. On average each graph has 38.7 nodes and 48.3 edges, and each text article contains 3,533.8 tokens.
In addition to structural information, our graphs also contain rich text information with an average of 895.1 tokens in each graph.  Furthermore, the automatic process we used to create this dataset can be extended to pair any Wikipedia document with Freebase, and can be scaled up to create over 3M graph-text pairs.

Out of many exciting new tasks that this dataset enables, we present 3 possibilities: graph $\rightarrow$ text generation, graph $\rightarrow$ text retrieval, and text $\rightarrow$ graph retrieval.  We benchmarked a few baseline models on these tasks.  The models we considered were based on the recent Transformer-XL \cite{dai2019transformer} model, and we adapted it to condition the text generation on the KG in different ways.  Our results show that better conditioning on the graph indeed improves the relevance of the generated text and the retrieval quality.  However, there is still significant room for improvement on these tasks, which makes this an exciting dataset for research.
Our data and code for baseline models will be made publicly available. 
 




\section{Related work}

\paragraph{Graph-text paired data} There has been a lot of prior work on creating graph-text paired datasets.
Example applications include generating text summaries conditioned on Abstract Meaning Representation graphs \cite{liu2018toward}, generating the abstract of a scientific article given a KG and title \cite{koncel-kedziorski-etal-2019-text} and generating text from RDF triples \cite{gardent-etal-2017-webnlg,jin-etal-2020-genwiki}.  In the following we will mostly review related work on KG - text paired datasets.

Annotating KG or text to create paired datasets is expensive, as
a good quality annotation requires annotators that understand the content and structure of the text and the corresponding KG \cite{jin-etal-2020-genwiki}.
Therefore previous KG-text paired datasets that rely on human annotation have limited scale.
Among these, \citet{gardent-etal-2017-webnlg} crowdsourced human annotators to verbalize RDF triplets taken from DBpedia \cite{DBpedia} to a few sentences (WebNLG) and this caused errors in annotation that were fixed with a few updates through years.
\citet{parikh2020totto} paired Wikipedia Table with one sentence text that is created by annotators that revise Wikipedia text.

Another line of research focuses on eliminating the need of human annotations by automatically matching KG-text pairs or generating KGs from text using existing tools. 
\citet{lebret-etal-2016-neural} automatically matched Wikipedia infobox of biographies with their first sentence.
\citet{koncel-kedziorski-etal-2019-text} utilized an earlier information extraction system that extracts entities, co-reference and relations from given text to build KG's. 
The GenWiki dataset \cite{jin-etal-2020-genwiki} is automatically constructed by querying KGs in DBpedia with the title of articles in Wikipedia followed by filtering and entity annotation. 


\begin{table}[t]
    \centering
    \setlength{\tabcolsep}{3pt}
    \begin{tabular}{c|c|c|c|c}
    \hline
        Dataset & \#examples & \#triples & \#tokens & \#vocab \\
        \hline
        WebNLG &13,036 & 2.54 & 15.26 & 1,484 \\
        GenWiki & \bf{1.3M} & 1.95 & 21.46 & \bf{476,341} \\
        Ours & 23,522 & \bf{48.3} & \bf{3,533.8} & 238,071\\
        \hline
    \end{tabular}
    \caption{Our dataset contains significantly larger graphs (average \#triples per graph) and longer text (average \#tokens per text) than previous KG-text datasets. 
    }
    \label{tab:dataset-comparison}
\end{table}

We construct our \ourdataset~dataset by extracting a subgraph from Freebase \cite{bollacker2008freebase} for each Wikipedia article following a scalable automatic process.  Compared to previous work, our \ourdataset~dataset contains significantly larger graphs and longer text (\tabref{tab:dataset-comparison}).

\paragraph{Models for graph-text paired data}
Recent state of art language models are based on the Transformer architecture \cite{Vaswani2017} that uses the
self attention mechanism.
The Transformer-XL \cite{dai2019transformer} model further introduces a segment level recurrence with a novel positional encoding resulting in impressive performance in long sequences by capturing dependencies beyond a fixed length window.

Graph neural networks (GNNs) \cite{battaglia2018relational,gilmer2017neural} learn representations for graph structured data through a message passing process. This class of models naturally exploit the graph structures, 
making them a good fit for graph data.  GNNs have been used in many applications on 
KG's \cite{kipf2016semi,wang2019knowledge,xu2019cross}.  Fundamentally, transformers can also be understood as a special type of GNNs with a fully-connected graph structure.

  
The most recent prior work on graph-to-text generation
follows an encoder-decoder architecture \cite{koncel-kedziorski-etal-2019-text, jin-etal-2020-genwiki}, where the graph part is encoded with a GNN model, e.g. 
Graph Attention Network (GAT) \cite{velickovic2018graph}.
The text part is typically modeled using an attention based decoder with a copy mechanism (e.g. BiLSTMs as in \cite{jin-etal-2020-genwiki}) to process input from both the KG and text.

The models we benchmarked for graph-to-text generation were based on the Transformer-XL architecture and conditioned on the graph through a GNN, making full use of the graph structure and capable of generating very long text comparable to the state-of-the-art.

\section{Dataset}

In this section we first present some properties of our dataset, and then describe the process that we used to create it.

\subsection{Properties of the data}

\begin{table}[t]
\centering
\setlength{\tabcolsep}{3.5pt}
{\small
\begin{tabular}{c|c|c|c|c}
\hline
    & Train & Valid & Test & All\\
    \hline
    Num. pairs & 23,431 & 48 & 43 & 23,522\\
    \% of WikiText-103 & 82.3\% & 80.0\% & 71.7\% & 82.3\%\\
    Nodes per graph & 38.7 & 35.4 & 40.6 & 38.7 \\
    Edges per graph & 48.3 & 42.8 & 49.5 & 48.3 \\
    Avg. Node degree & 2.5 & 2.4 & 2.4 & 2.5 \\
    Tokens per graph & 895.1 & 807.7 & 1,010.1 & 895.1 \\
    Total graph tokens & 21.0M & 38,771 & 43,435 & 21.1M \\
    Graph vocab size & - & - & - & 31,090 \\
    Tokens per article & 3,531.7 & 3,644.2 & 4,564.7 & 3,533.8 \\
    Total text tokens & 82.8M & 174,923 & 196,280 & 83.1M \\
    Text vocab size & - & - & - & 238,071 \\
    \hline
\end{tabular}
}
\caption{Basic statistics about our \ourdataset~dataset. 
}
\label{tab:basic-stats}
\end{table}


\subsubsection{Scale of the data}

Basic statistics about our \ourdataset~dataset are listed in \tabref{tab:basic-stats}.  An illustration of a graph-text pair is shown in \figref{fig:cartoon-example}.  A few actual examples from our dataset are included in the Appendix (\figref{fig:graph-1}, \ref{fig:graph-2}).  All of the articles come from the WikiText-103 dataset \cite{merity2016pointer}, which contains high-quality articles that fit the \emph{Good} or \emph{Featured} criteria specified by the Wikipedia editors when the data was collected.  \citet{merity2016pointer} have already cleaned up and tokenized the articles, therefore they appear as plain text without any markup tags.

As will be described in \secref{sec:data-creation}, we try to pair each article with a subgraph from Freebase, centered at the entity node that has a Wikipedia link to the title of the article.  We are not able to match every article to an entity in Freebase, but through this process we retained a significant portion of 82.3\% of the WikiText-103 articles.  We kept the original train/valid/test split.  As we will see in \secref{sec:graph2text}, training models on this set gives us results that are very close to training on the full WikiText-103 dataset when evaluated on our test set.
Therefore the text part of \ourdataset~appears to be sufficient to reproduce and benchmark against the state-of-the-art text generative models. 

\begin{figure*}[t]
    \centering
    \includegraphics[width=1.0\textwidth]{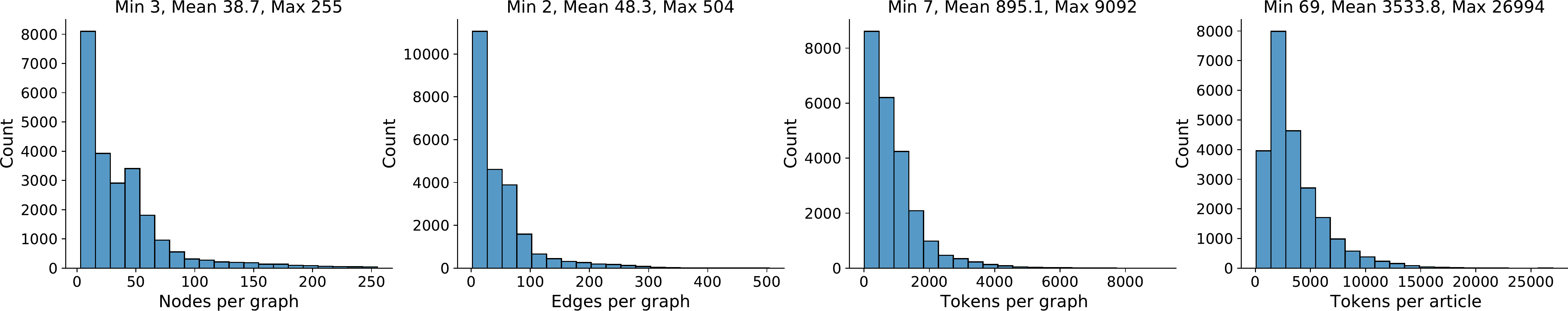}
    \caption{Distribution of graph and article sizes across our \ourdataset~dataset.}
    \label{fig:dataset-stats}
\end{figure*}

\figref{fig:dataset-stats} shows the distribution of graph sizes and article lengths across our dataset.  All the distributions are skewed with a long tail.  Notably, average graph size in our dataset is 38.7 nodes and 48.3 edges, considerably larger than the graphs in previous datasets \cite{jin-etal-2020-genwiki, gardent-etal-2017-webnlg}. 
Also the length of the text articles averages to 3,533.8 tokens and can go up to 26,994 tokens, which is orders of magnitudes longer than the text data in previous graph-text paired datasets that typically only contains a single or few sentences \cite{jin-etal-2020-genwiki,gardent-etal-2017-webnlg, lebret-etal-2016-neural}.

\subsubsection{Nodes and edges}

\begin{figure}[th]
    \centering
    \includegraphics[width=0.7\columnwidth]{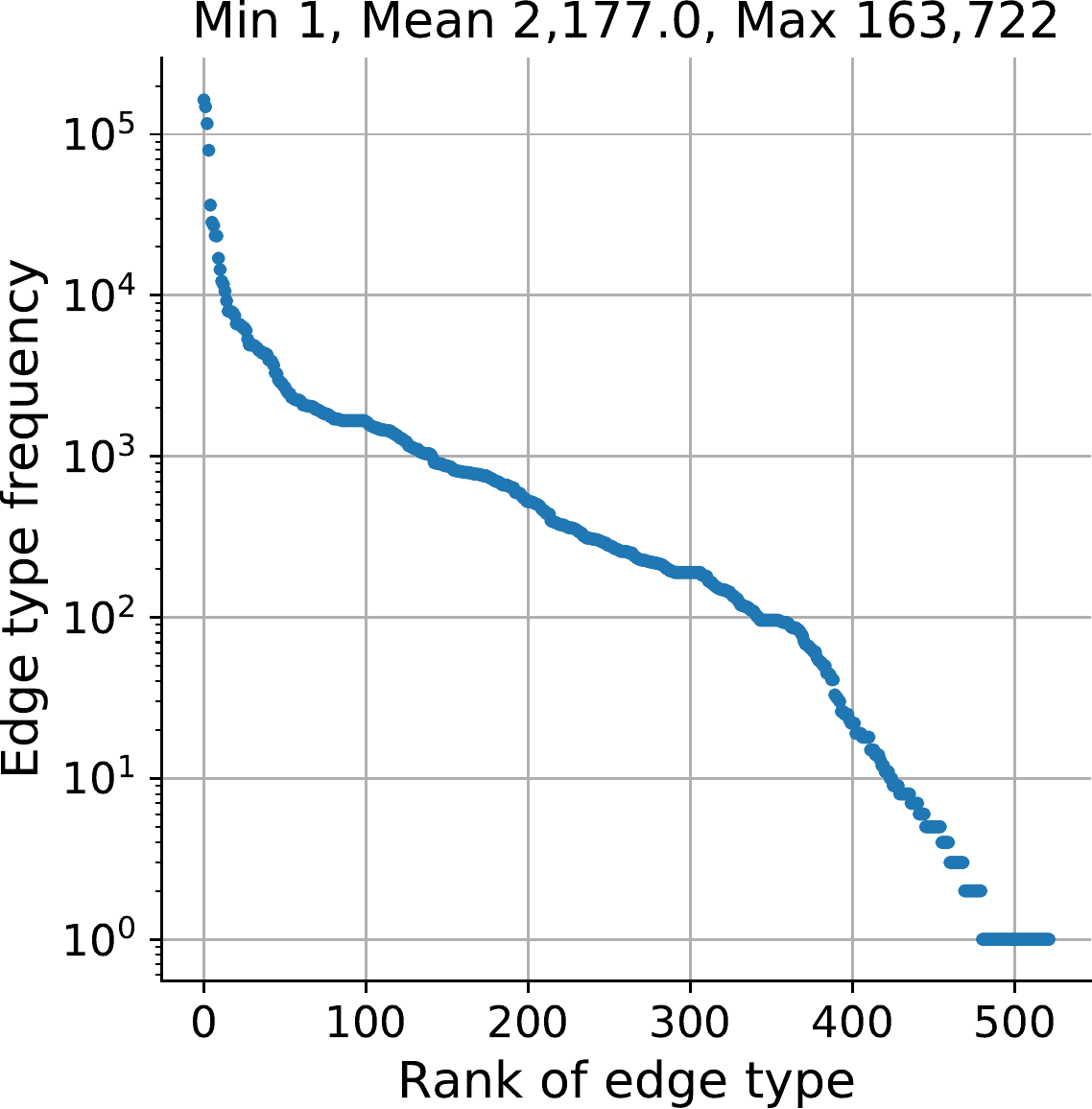}
    \caption{Edge type distribution roughly follows an inverse exponential law.}
    \label{fig:edge-type-dist}
\end{figure}

The graphs in our dataset contains two types of nodes: entities and string literals.  Each entity is labeled by a unique Freebase entity ID, e.g. \texttt{ns/m.0f9q9z}, and each string literal contains some natural language text, that could be for example a name, date, or description of an entity.  Each edge in the graphs also has an associated edge label, e.g. \texttt{ns/common.topic.description}, indicating which type of edge it is.  There are a total of 522 different edge types in our dataset.  \figref{fig:edge-type-dist} shows the frequency of all the different edge types in our dataset.

Every graph always has one entity node (we call it ``center node'') that has a link to the paired Wikipedia article, through a special edge \texttt{key/wikipedia.en}, and the whole graph is a 1-hop neighborhood of entities around the center node within the bigger Freebase KG, plus the string literals associated with all the entities included.  Note that it is possible to have edges between the 1-hop neighbors of the center node, therefore the graphs typically are not star structured. \secref{sec:data-creation} provides more details about how these graphs are constructed and any additional filtering we did.

\begin{figure}[t]
    \centering
    \includegraphics[width=0.8\columnwidth]{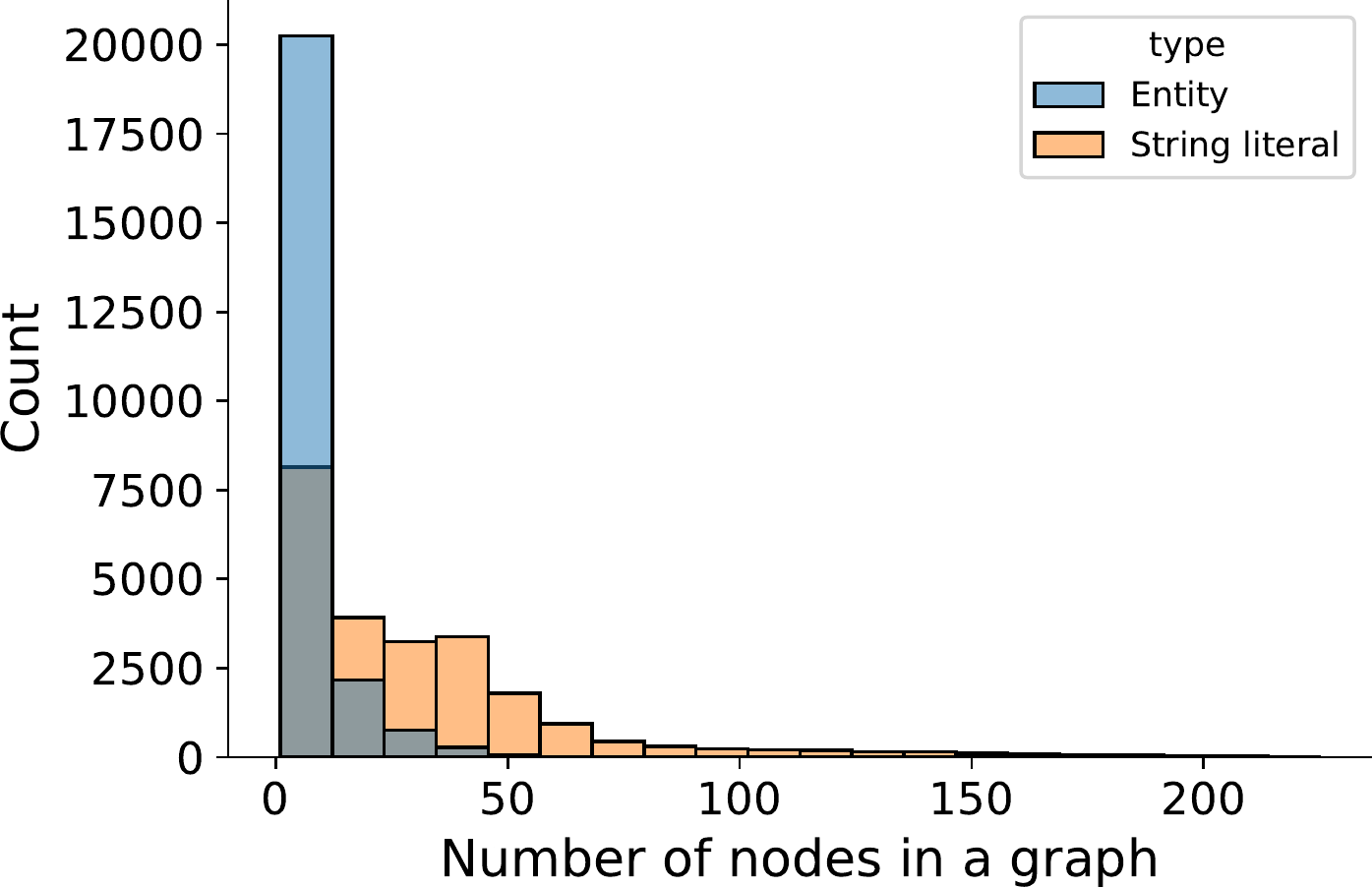}
    \caption{Distribution of the per-graph number of entity nodes and string literal nodes in our dataset.}
    \label{fig:node-type-stats}
\end{figure}

One special characteristic about our graph data is that the natural language text contained in the string literal nodes can sometimes be quite long (see e.g. \figref{fig:graph-1},\ref{fig:graph-2}), and therefore provide much richer information not included in the graph structure itself.  On average, each graph contains 895.1 tokens across all the string literal nodes in one graph (\tabref{tab:basic-stats}, \figref{fig:dataset-stats}, ``Tokens per graph'').

\figref{fig:node-type-stats} shows the distribution of per-graph number of entity nodes and string literal nodes in our dataset.  We can see that our graphs tend to have more string literal nodes than entity nodes, indicating that the entities are supplemented with the rich information in the string literals.

\begin{figure*}[ht]
    \centering
    \begin{tabular}{cccc}
        \includegraphics[width=0.3\textwidth]{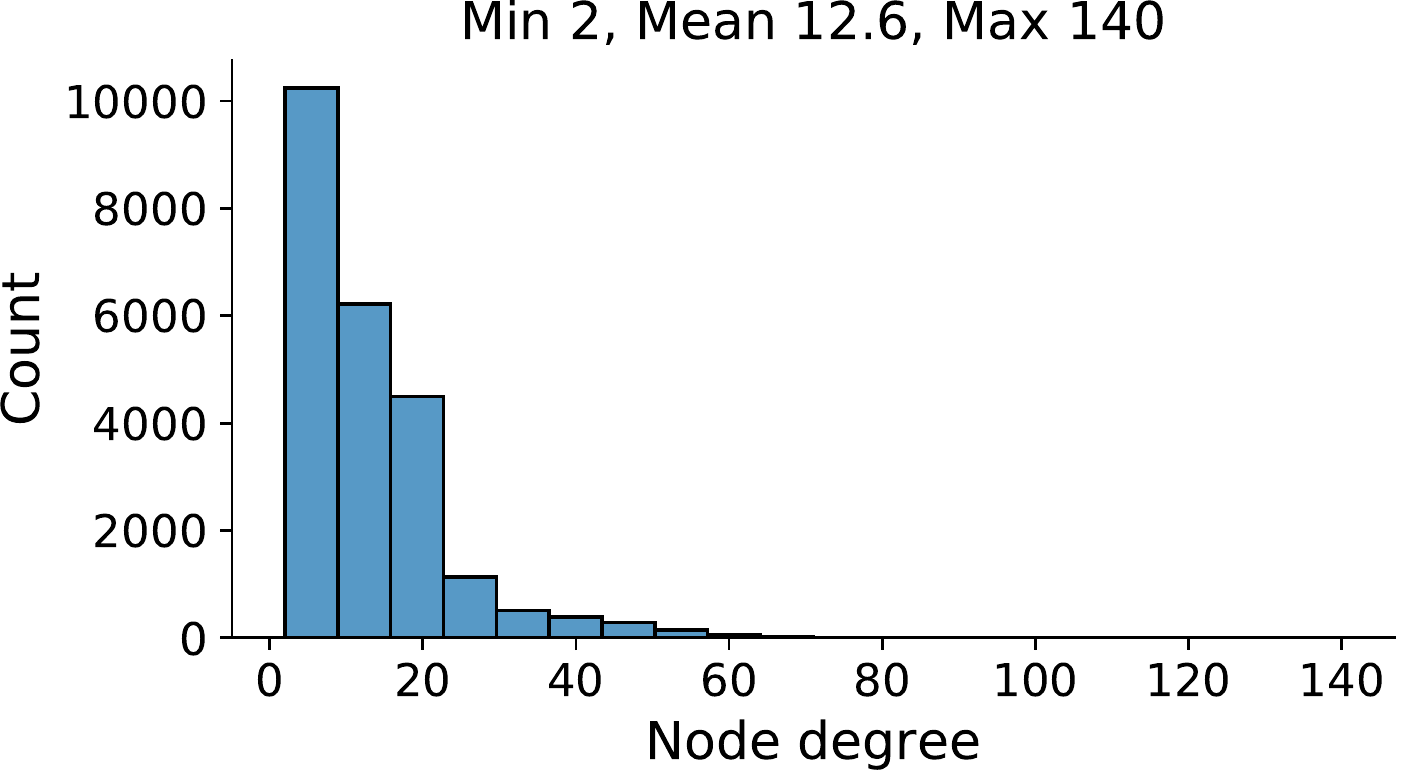} &
        \includegraphics[width=0.3\textwidth]{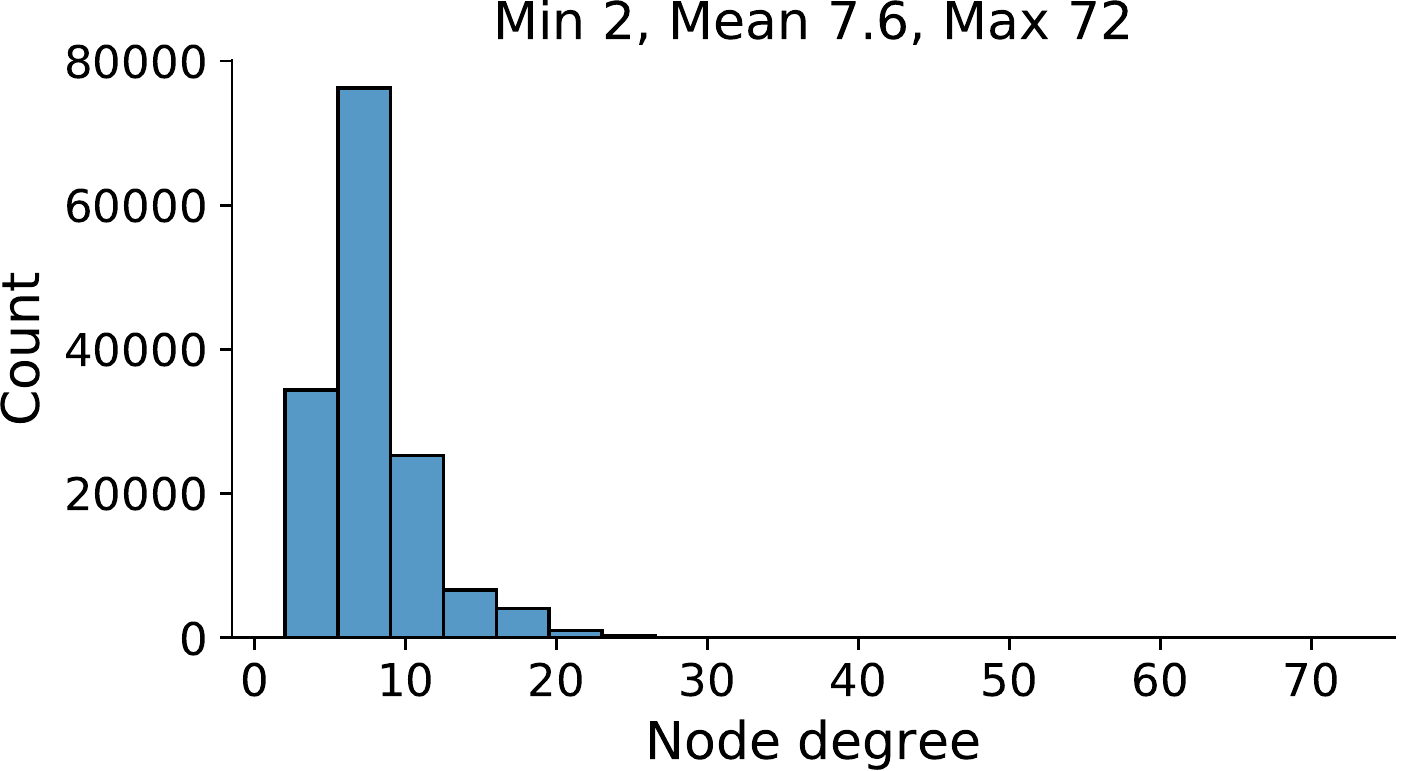} & 
        \includegraphics[width=0.3\textwidth]{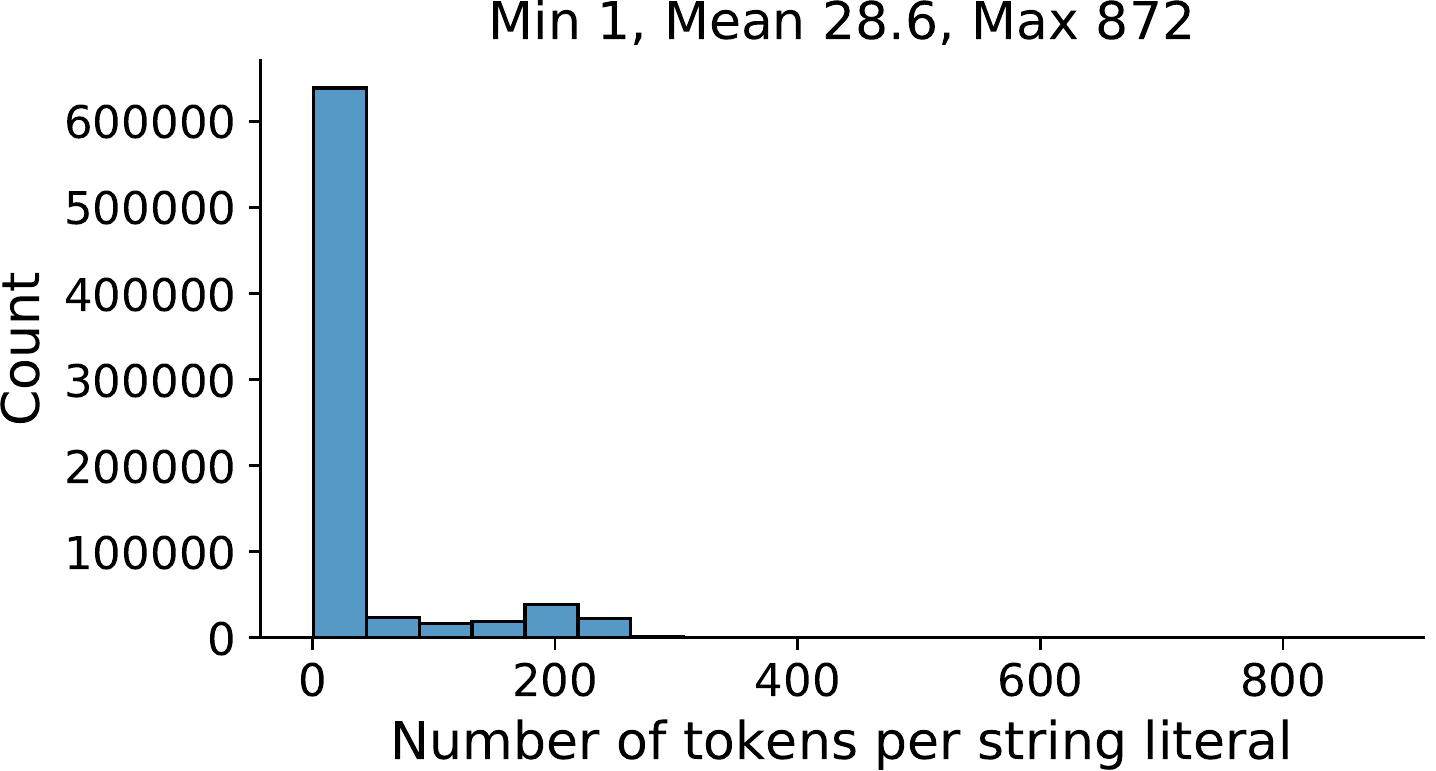} \\
        (a) Center node degree dist. &
        (b) Non-center node degree dist. &
        (c) String literal node length dist.
    \end{tabular}
    \caption{Node degree distribution for entity nodes and  token count distribution for string literal nodes.}
    \label{fig:node-dist}
\end{figure*}

The distribution of information is not uniform across the nodes in a graph.  \figref{fig:node-dist} shows that most entity nodes in our graph has a small degree, while few nodes have much larger degrees.  Also most string literal nodes contain short text, while fewer nodes contain longer text.

The skewed distribution of nodes and edges in our dataset reflect the nature of KG's like Freebase, and presents new challenges to graph representation learning models.

\subsection{The dataset construction process}\label{sec:data-creation}

We follow three principles when designing the dataset construction process:
\begin{enumerate}
    \item\label{enum:text-principle} The text part of the data should be directly comparable in complexity to the capability of state-of-the-art text generative models.
    \item\label{enum:graph-principle} The graph part of the data should be constructed in an automatic and scalable way.
    \item\label{enum:relevance-principle} The graph part of the data should be relevant for the paired text data.
\end{enumerate}

Note that our process is general, and can be applied to any set of Wikipedia articles.  We have tried to pair a full dump of English Wikipedia with Freebase and managed to get over 3 million graph-text pairs.  Here we restrict the process to the set of articles from the WikiText-103 dataset.

We try to map each Wikipedia article
to a relevant subgraph of the existing large scale KG Freebase \cite{bollacker2008freebase}.
We used the last public dump of Freebase\footnote{\url{https://developers.google.com/freebase}}, which contains 1.9B triples and a total of 250GB of data. We filtered the data by keeping only the entities with at least 4 string attributes (otherwise the entities are less interpretable), and keeping only the top 1024 most frequent relation types and restricting the relations to only those among the retained entities and between the entities and string attributes.  We also simplified the entity and relation names by stripping off the irrelevant ``http://rdf.freebase.com/'' and further removed duplicates. This gives us a significantly cleaner and smaller backbone graph for Freebase, with about 20M nodes.

Finding the relevant subgraph for an article in such a cleaned up but still large KG remains non-trivial.  Our process for this contains 3 stages: \emph{mapping}, \emph{expansion}, and \emph{filtering}.

\paragraph{Mapping}  In the first stage of the process, we map each article into an entity in our processed Freebase KG.  This is made possible through triples from Freebase like the following:
\begin{tcolorbox}
{\small
ns/g.11b6jbqpt4~~~~key/wikipedia.en~~~~"Madunnella"}
\end{tcolorbox}
\noindent
where ns/g.11b6jbqpt4 refers to an entity in the KG, key/wikipedia.en is the type of the edge, which indicates that this entity is linked to a Wikipedia article and ``Madunnella'' is the title of that article.  We normalize the title string (and in general any string literals) from Freebase by replacing ``\_'' with white space and handle unicode characters properly.  We extract the titles from the Wikipedia article through string matching, where titles are enclosed in a ``= [title] =" pattern.

In this step we managed to map 24,345 out of 28,475 (85.5 \%) article titles from WikiText-103 to an entity in our KG.

\paragraph{Expansion}  We treat each of the mapped entities as the center node of a subgraph, and expand 1 hop out in the entire filtered Freebase graph to include all the neighboring entities that are the most relevant to the center entity.  We then expand further from this 1-hop graph out to include all the relations that connect the selected entities to string attributes as well as between these entities themselves.  Note that because of these edges between the 1-hop neighbor entities the graphs are typically not star structured. This gives us a relevant but compact graph for each article.  We have also investigated the possibility of a 2-hop neighborhood from the center node, and found that 2-hop neighborhoods are significantly larger than 1-hop and through some ``hub'' nodes like ``Male'' or ``Female'' a 2-hop neighborhood from an entity can easily include many other irrelevant entities.  Based on such observations we decided to use the 1-hop neighborhood to keep the relevance of the subgraph high.

\paragraph{Filtering}  The last stage of the process involves more filtering and cleaning up of the data.  We noticed that in Freebase it is common for one entity to have multiple relations of the same type pointing to different string attributes, like the following:
\begin{tcolorbox}
{\small
ns/m.07c72~~~~key/wikipedia.en~~~~"The SImpsons" \\
ns/m.07c72~~~~key/wikipedia.en~~~~"The Simpson" \\
ns/m.07c72~~~~key/wikipedia.en~~~~"The simsons" \\
ns/m.07c72~~~~key/wikipedia.en~~~~"Thr Simpsons" \\
ns/m.07c72~~~~key/wikipedia.en~~~~"The Simpson's"}
\end{tcolorbox}
It is clear that there is a lot of redundancy in this data.  We reduced all such edges (from the same entity with the same edge type to string attributes) to a single edge by picking the most ``canonical'' one.  This was done by fitting a unigram model to the characters in the collection of strings and using that model to pick the most likely string.

We also filtered the graphs based on size and created three versions of the data with maximum graph size capped at 256, 512, and 1024 nodes, respectively.  All the statistics and results in the rest of the paper are based on graphs with a maximum size of 256, but all versions of the data are made available online. 

\section{Experiments}

We perform a set of experiments to showcase how the text and graph information can be combined in a language model. Specifically, we consider three tasks: text generation conditioned on the graph, graph retrieval given the text, and text retrieval given the graph.

\subsection{Graph-conditioned Transformer-XL}
In order to incorporate graph information into an advanced language model, we adapt the recent Transformer-XL model \cite{dai2019transformer} to also attend to the graph features.  At a high-level our model embeds the graph into a set of embedding vectors, and then exposes these embeddings to the Transformer-XL model as extra ``token'' embeddings to condition on.  The size of this set depends on the graph model we choose. 

Given the features for $T$ text tokens $\mathbf{H}_t \in \mathbb{R}^{T \times d}$ and features for $T'$ graph ``tokens'' $\mathbf{H}_g \in \mathbb{R}^{T' \times d'}$, we illustrate the graph-conditioned attention procedure with a single head as follows:
%
\begin{align*}\label{eq1}
  \mathbf{Q}_t, \mathbf{K}_t, \mathbf{V}_t &= \mathbf{H}_t\mathbf{W}^t_q, \mathbf{H}_t\mathbf{W}^t_k, \mathbf{H}_t\mathbf{W}^t_v \\
  \mathbf{K}_g, \mathbf{V}_g &= \mathbf{H}_g\mathbf{W}^g_k, \mathbf{H}_g\mathbf{W}^g_v \\
  \mathbf{A}_t, \mathbf{A}_g & = \mathbf{Q}_t\mathbf{K}_t^\top,  \mathbf{Q}_t \mathbf{K}_g^\top \\
  \mathbf{A}, \mathbf{V} & = [\mathbf{A}_t \circ  \mathbf{A}_g], [\mathbf{V}_t \circ  \mathbf{V}_g] \\
  \mathbf{O} & = \textup{Masked-Softmax}(\mathbf{A}) \mathbf{V}
\end{align*}
where $[a \circ b]$ stands for concatenation on the sequence dimension and thus $\mathbf{A} \in \mathbb{R}^{T \times (T + T')}$ and $\mathbf{V} \in \mathbb{R}^{(T + T') \times d_h }$, where $d_h$ is the head dimension.
In other words, comparing to the original Transformer-XL, our model also computes the attention scores between the text queries $\mathbf{Q}_t$ and both the text keys $\mathbf{K}_t$ and the graph keys $\mathbf{K}_g$.
As a result, the attention outputs contain information from both the graph and the text context. Note that this formulation is compatible with an additional memory \cite{dai2019transformer} with minimal changes, as it simply adds in an extra set of ``tokens'' for the model to attend to.  We don't use position encodings for the graph ``tokens'' as there is no sequential ordering for them.

In this work we consider three different approaches for encoding the graph structure:
\begin{itemize}
    \item \textbf{Bag-of-words (BoW):}
    we construct a single bag-of-words representation of all the tokens from both the nodes and edges in the graph.
    Entity IDs and numeric values in the graph are replaced with special tokens \texttt{<entity>} and \texttt{<number>}. The BoW vector is further projected using a linear layer to a latent space. In this case $T'=1$.
    \item \textbf{Nodes only (Nodes)}: we construct separate BoW representations for each node and project each to an embedding and ignore the edges. In this case $T'$ is equal to the number of nodes in the graph.
    \item \textbf{Graph neural network (GNN):} we embed BoW representations for both nodes and edges and then use a graph neural network  \cite{battaglia2018relational} on top of those embeddings to compute a new set of node embeddings. $T'$ is equal to the number of nodes.
\end{itemize}
The $T'$ graph embeddings from this process are shared across all the time steps for text tokens.
This model can be further improved, e.g. by using word embeddings and text summarization techniques, but we leave these for future work.

\subsubsection{Implementation details}
We reimplement the Transformer-XL model in Jax \cite{jax2018github}. In our experiments, we employ the base model in \cite{dai2019transformer}, except that we increase the tail shrinkage factor used for the adaptive softmax and input representations from 1 to 4, which saves $63\%$ of the parameters without compromising the performance. On the full Wikitext-103 dataset, our implementation has a test perplexity of 24.2 (published result for this base model was 24.0). We train our models using the standard likelihood objective for language models with a total batch size of 64 on 8 V100 GPUs. Adam optimizer is used with an initial learning rate of $2.5\times10^{-4}$, which decays up to 200k steps following a cosine curve. During training, we use text segments of 150 steps and a memory of equal size. When evaluating the model, we use a sequence length of 64 and memory size 640. Unless further noted, in our experiments we use an embedding size of 256 for BoW-conditioned models. For other models, we project each node or edge represented by BoW to an embedding space of size 128. The default GNN we use has a single linear message passing layer of 256 hidden units.



\subsection{Graph $\rightarrow$ text generation}\label{sec:graph2text}


Our first task is text generation conditioned on the graph.  We evaluate model performance by (1) computing model perplexity on held-out text and (2) drawing samples from the model and comparing that to the ground truth text article.  We use BLEU score \cite{papineni2002bleu} to measure the similarity of our generated samples to the ground truth.  Unlike previous use cases for BLEU score where there are many references for one generated sample, here we have only one ground truth reference but we can generate multiple samples.  We therefore simply swapped the reference with the samples when computing the score, which we term as the reverse-BLEU (rBLEU).  We have also tried other ways of computing the BLEU score and find that they don't change how models compare against each other.


Unless explicitly stated, we let the model sample with a memory size of 640, and condition on the graphs in the test set to generate text for up to 512 tokens per sample for a total of 20 samples per graph.
The rBLEU score is computed based on these samples and corresponding ground-truth texts are truncated to the same length.  We sample the texts from the distribution with a temperature of 0.8.  For each case, we report the average rBLEU score of 3 sampling runs. We find the variances are insignificant which do not affect the comparison results. In Appendix~\ref{sec:appendix_sampling} we also report results for generating longer samples for up to 4096 tokens.

\subsubsection{Main result}


\begin{table}[t]
    \centering
    \setlength{\tabcolsep}{5pt}
    \begin{tabular}{c|c|c|c|c|c}
    \hline
        \multirow{2}{2em}{Cond.} & \multirow{2}{2em}{Test Ppl.} & \multicolumn{2}{c|}{rBLEU} & \multicolumn{2}{c}{rBLEU(w/title)} \\ \cline{3-6}
          & & Valid & Test & Valid & Test \\
        \hline
        None & \textbf{25.85} & 10.97 & 9.98 & 27.98 & 24.07 \\
        BoW & 26.65 & 29.53 & 24.41 & 32.41 & 27.39  \\
        Nodes & 27.40 & 30.51 & 25.31 & 32.60 & 27.43 \\
        GNN & 26.93 & \textbf{31.39} & \textbf{26.22} & \textbf{32.65} & \textbf{28.35} \\
        \hline
    \end{tabular}
    \caption{The perplexity and the generated text reverse-BLEU score of different types of graph-conditioned models. We show the reverse-BLEU score with or without prompting the original title at the start of the text generation.}
    \label{tab:graph2text}
\end{table}

In Table~\ref{tab:graph2text}, we show the perplexity and the rBLEU score of the unconditional, BoW, nodes-only, and GNN conditioned models.  As a reference, a standard Transformer-XL model trained on the full Wikitext-103 training set reaches 25.08 perplexity on our test set, which contains 71.7\% of the original test articles. We can see that the unconditional, i.e. text only, model trained on our dataset gets a very similar performance as trained on the full set.  This is strong evidence that our dataset can be a good benchmark for state-of-the-art text generative models.

We also see that conditioned on the graphs, model perplexity didn't improve, but the relevance of the samples measured by the BLEU scores did improve significantly.  This indicates that the graph conditioned models can indeed steer the language model towards more relevant topics, but this so far cannot yet improve likelihood metrics. 

To make the evaluation more fair to the text-only model, we also tried to prompt the generation with the title of the article, such that the text-only model also has some context.  In this setting the graph models are still better, showing the importance of modeling the structure.

Lastly, among all the 3 graph model variants, we observe that using a set of embeddings from the nodes model is better than using a single embedding from the BoW model, and fully utilizing the graph structure through the GNN model is consistently better than ignoring the edges as in the nodes model.  However the differences among the methods are relatively small.
For visualizations of a few graphs in our dataset and the corresponding samples generated based on them please refer to Appendix~\ref{sec:appendix}.



\begin{table}[t]
    \centering
    \begin{tabular}{c|c|c}
    \hline
        \# MP layers & Test Ppl. & Test rBLEU  \\
        \hline
        0 & 26.65 & 25.31 \\
        1 & 27.40 & 26.22  \\
        3 & 27.20 & 26.16 \\
        5 & 26.85 & 25.91 \\
        \hline
    \end{tabular}
    \caption{The test perplexity and the generated text reverse-BLEU score (without title prompt) of GNN-based models with different numbers of message passing layers.}
    \label{tab:gnn_layers}
\end{table}
\begin{figure}[t]
    \centering
    \includegraphics[width=0.9\columnwidth]{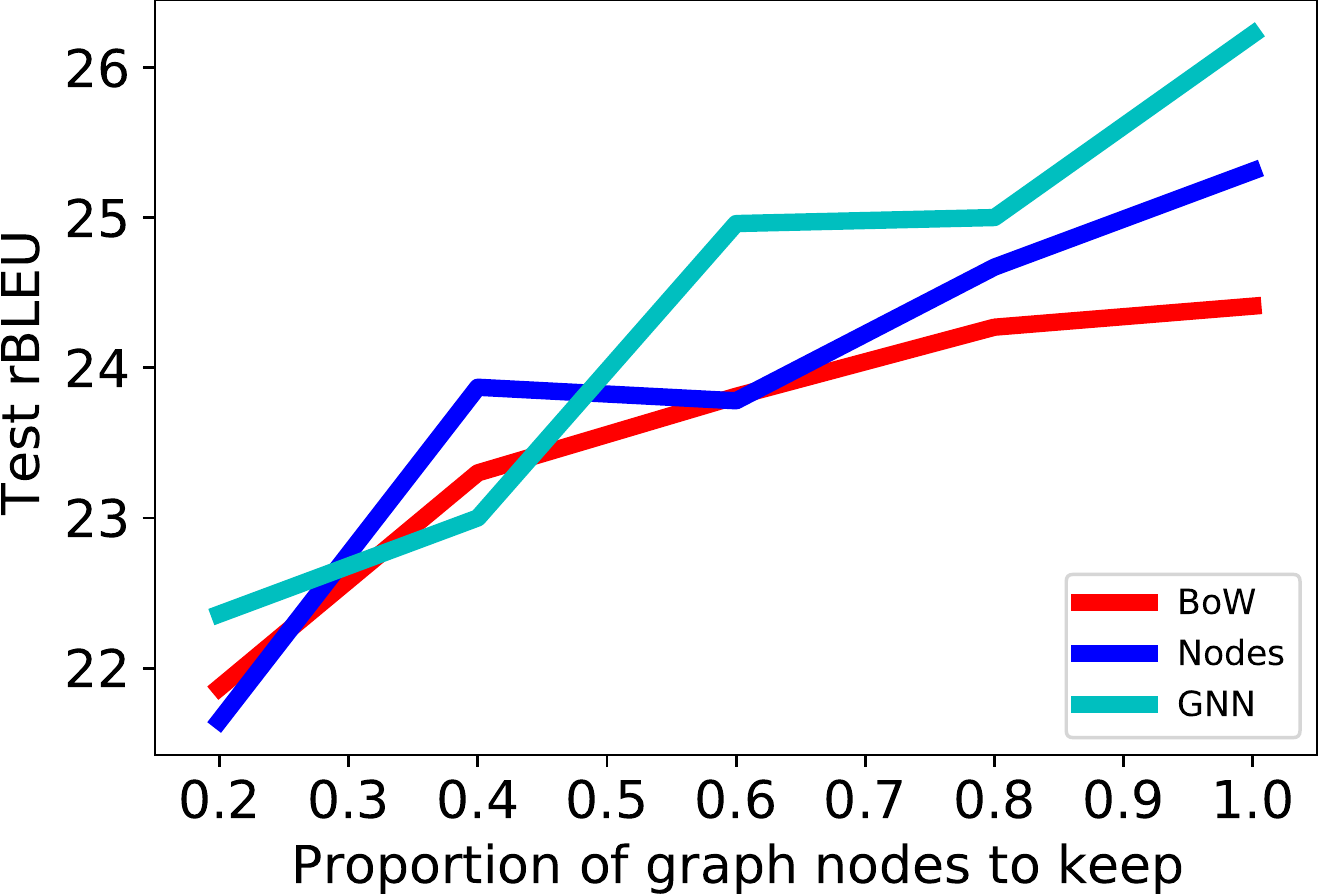}
    \caption{Performance vs size of graph to condition on. The model is trained with a smaller version of the data by subsampling the number of nodes.
    }
    \label{fig:graphsize}
\end{figure}



\subsubsection{Ablation studies}

We show a few ablations on the graph model and sampling parameters, to provide some insights into the models.

\tabref{tab:gnn_layers} shows the effect of varying the number of message passing layers in the GNN. 
We can observe that there is a big difference between using message passing ( $\ge$ 1 layers) or not (0 layers) in terms of rBLEU score, but increasing the number of message passing layers does not change the results significantly.  
We believe however, that these results can be improved by employing bigger and more powerful graph representation learning models, and potentially use initial node and edge representations better than bag-of-words.

In \figref{fig:graphsize} we show the effect of the graph size on model performance.  In this experiment we subsample the nodes in each graph to control for the amount of context the model has access to.  It is clear from the results that when we heavily subsample and keep only a small portion of the graphs, the GNN model performs similarly as the simpler BoW model, but GNNs benefit more as we keep more of the graph structure.

\subsection{Graph $\rightarrow$ text retrieval}
\label{sec:g2t_retrieval}
In this task, we evaluate the possibility of retrieving relevant text for a given query graph.  We pair all articles with all graphs in the test set, resulting in 43$\times$43$=$1849 pairs.
Then the trained graph-conditioned language models are used to produce the per-token likelihood of each pair, and we use these likelihood scores to rank the text articles for each graph.  We expect the learned models can rank the correct pairs higher than wrong ones. 
To measure the results we use standard ranking metrics including recall@K, which computes the fraction of times the correct pair is included in the top K predictions, as well as mean average precision (mAP). In Table~\ref{tab:graph2text-retrieval}, it is observed that graph-conditioned models can indeed retrieve more relevant texts from the graph than the unconditional model,
among which the GNN-based model performs the best, and the unconditional model performs close to a random guess.

\begin{table}[t]
    \centering
    \begin{tabular}{c|c|c|c}
    \hline
        Cond. & Recall@1 & Recall@5 & mAP \\
        \hline
        None & 0.02 & 0.12 & 0.10 \\
        BoW & 16.28 & 30.23 & 25.98 \\
        Nodes & 16.28 & \textbf{34.88} & 26.62 \\
        GNN & \textbf{18.60} & \textbf{34.88} & \textbf{27.79} \\
        \hline
    \end{tabular}
    \caption{Text retrieval given the graph.
    }
    \label{tab:graph2text-retrieval}
\end{table}

\subsection{Text $\rightarrow$ graph retrieval}
In this last task, we evaluate the performance of graph retrieval given a text query.  We use exactly the same setting and scores as \secref{sec:g2t_retrieval}, but instead rank the graphs for each text article using the likelihood scores.
The results are shown in Table~\ref{tab:text2graph}. Note that this task is quite easy with our data and setup, potentially because the graphs are much more distinguishable than the text articles. 
All the graph-conditioned models perform almost perfectly, with the GNN model again outperforming the others.

\begin{table}[t]
    \centering
    \begin{tabular}{c|c|c|c}
    \hline
        Cond. & Recall@1 & Recall@5 & mAP \\
        \hline
        None & 0.02 & 0.07 & 0.02 \\
        BoW & 95.35 & \textbf{100.00} & 97.67 \\
        Nodes & 93.02 & \textbf{100.00} & 96.51 \\
        GNN & \textbf{100.00} & \textbf{100.00} & \textbf{100.00} \\
        \hline
    \end{tabular}
    \caption{Graph Retrieval given the text.}
    \label{tab:text2graph}
\end{table}

\section{Conclusion}

In this paper, we present WikiGraphs, a new graph-text paired dataset with significantly larger graphs and longer text compared to previous datasets of similar nature.  We show that the text part of this data is a good benchmark for state-of-the-art text generation models, and the paired dataset can help us benchmark models that are capable of generating long and coherent text conditioned on a graph structure.

In the first set of experiments on this dataset we showcase 3 different tasks using our dataset, and demonstrate the benefit of better models that make more use of the graph structure.

There is still significant room for improvement for these tasks on our dataset, and we hope the release of the data and baseline code can help spur more interest in developing models that can generate long text conditioned on graphs, and generate graphs given text, which is another exciting direction our dataset enables but we did not explore, and eventually bridging the graph and text modalities.




\bibliography{anthology,custom}
\bibliographystyle{acl_natbib}

\newpage
\appendix

\section{Appendix}
\label{sec:appendix}

\subsection{Graph visualization}
\label{sec:graph_vis}

\begin{figure*}
    \centering
    \begin{tabular}{c}
    \includegraphics[width=1\textwidth]{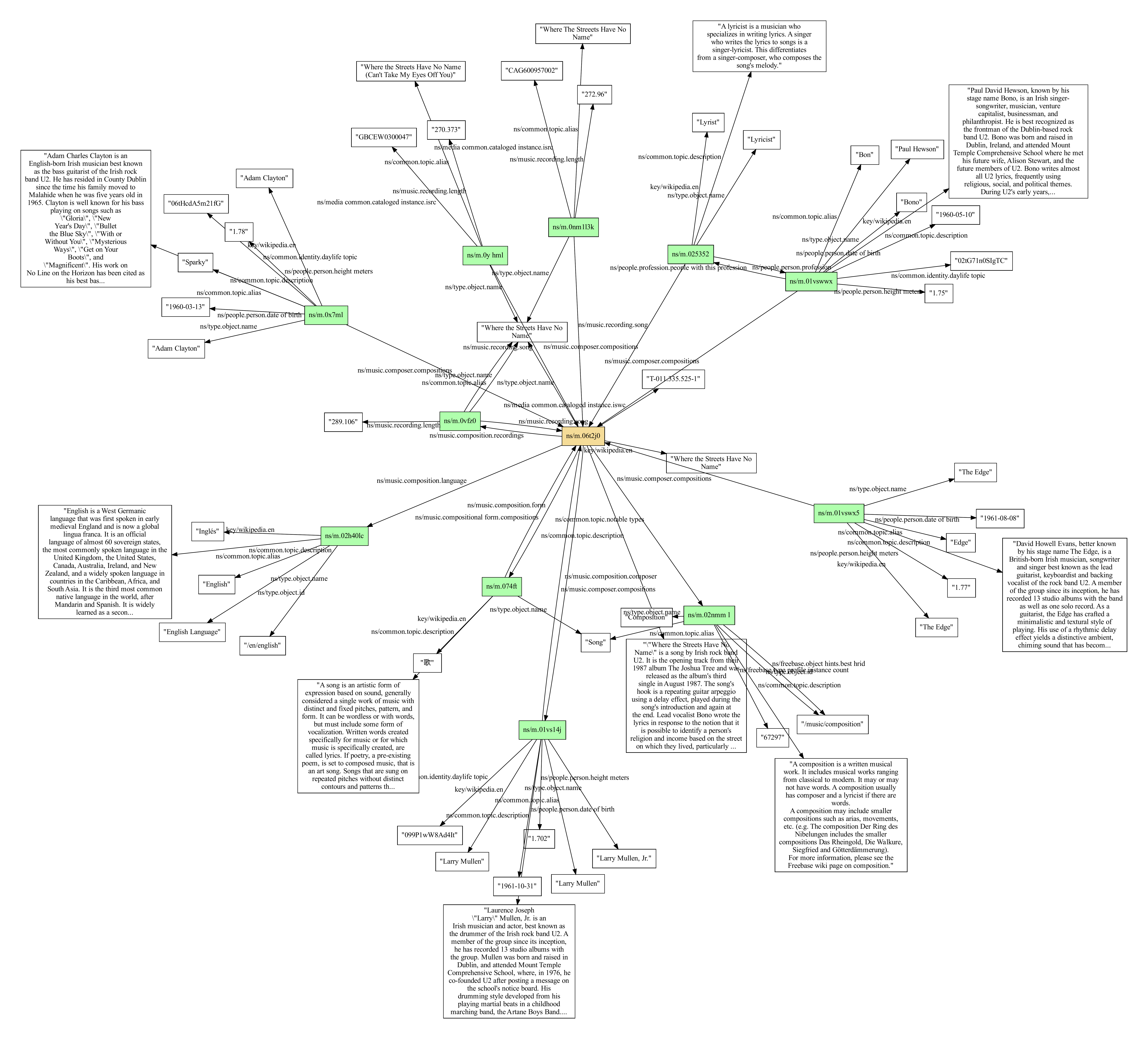} \\
    \end{tabular}
    \caption{Visualization of the ``Where the Streets Have No
Name'' KG in our dataset.
}
    \label{fig:graph-1}
\end{figure*}

\begin{figure*}
    \centering
    \begin{tabular}{c}
    \includegraphics[width=1\textwidth]{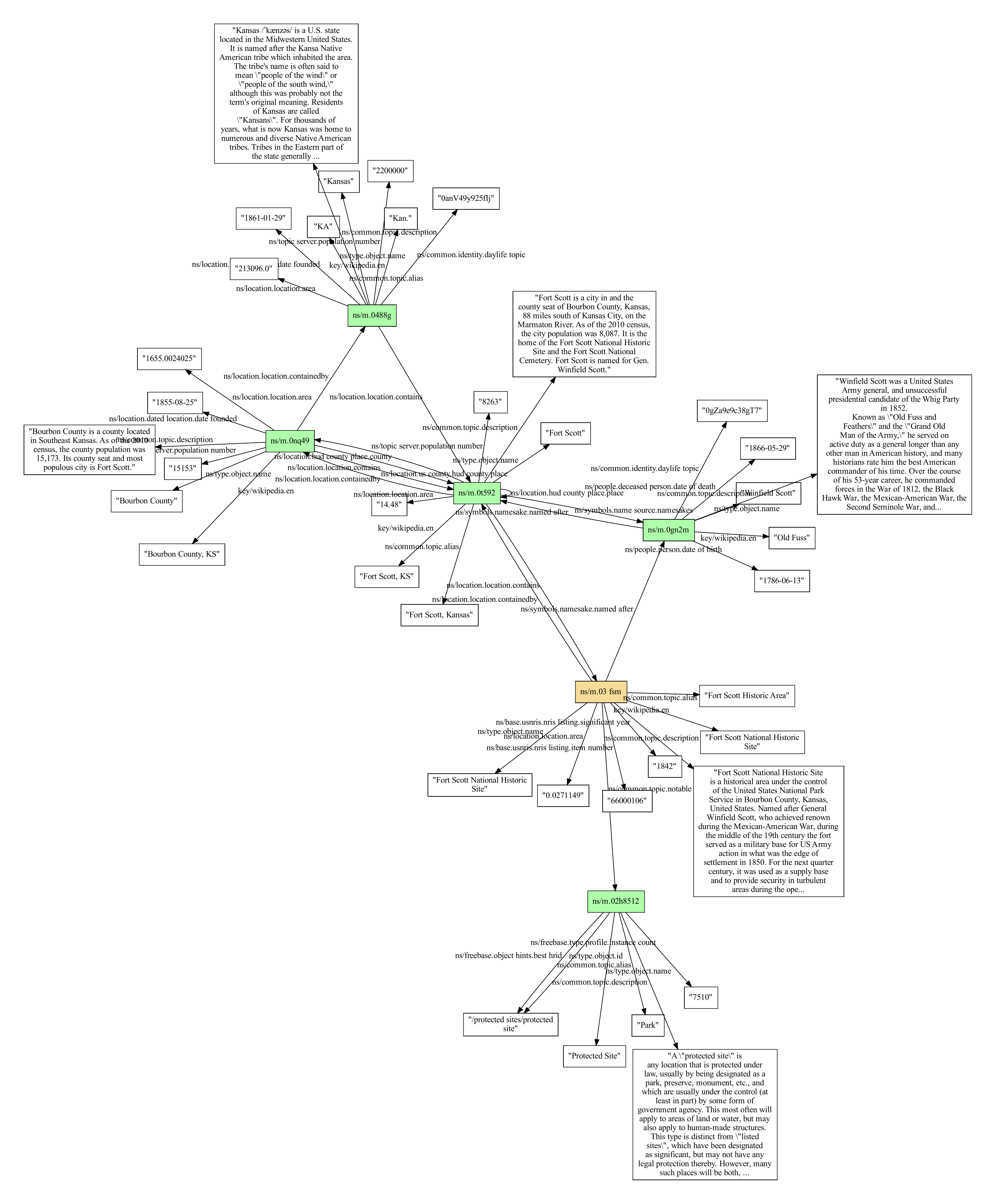} \\
    \end{tabular}
    \caption{Visualization of the ``Fort Scott National Historic Site'' KG in our dataset.}
    \label{fig:graph-2}
\end{figure*}

Some example visualizations of the KG structures are shown in \figref{fig:graph-1} and \figref{fig:graph-2}. The corresponding graph truth texts are shown in Table~\ref{tab:ground_truth_samples}.

\begin{table*}[h]
    \centering
    \begin{tabular}{p{0.15\linewidth} | p{0.8\linewidth}}
    \hline
        Visualization & Ground Truth Text  \\
        \hline
         Figure~\ref{fig:graph-1} & {\tiny = Where the Streets Have No Name = 
 
 " Where the Streets Have No Name " is a song by Irish rock band U2 . It is the opening track from their 1987 album The Joshua Tree and was released as the album 's third single in August 1987 . The song 's hook is a repeating guitar arpeggio using a delay effect , played during the song 's introduction and again at the end . Lead vocalist Bono wrote the lyrics in response to the notion that it is possible to identify a person 's religion and income based on the street on which they lived , particularly in Belfast . During the band 's difficulties recording the song , producer Brian Eno considered erasing the song 's tapes to have them start from scratch . 
 " Where the Streets Have No Name " was praised by critics and became a commercial success , peaking at number thirteen in the US , number fourteen in Canada , number ten in the Netherlands , and number four in the United Kingdom . The song has remained a staple of their live act since the song debuted in 1987 on The Joshua Tree Tour . The song was performed on a Los Angeles rooftop for the filming of its music video , which won a Grammy Award for Best Performance Music Video . 
 
 = = Writing and recording = = 
 
 The music for " Where the Streets Have No Name " originated from a demo that guitarist The Edge composed the night before the group resumed The Joshua Tree sessions . In an upstairs room at Melbeach House — his newly purchased home — The Edge used a four @-@ track tape machine to record an arrangement of keyboards , bass , guitar , and a drum machine . Realising that the album sessions were approaching the end and that the band were short on exceptional live songs , The Edge wanted to " conjure up the ultimate U2 live @-@ song " , so he imagined what he would like to hear at a future U2 show if he were a fan . After finishing the rough mix , he felt he had come up with " the most amazing guitar part and song of [ his ] life " . With no one in the house to share the demo with , The Edge recalls dancing around and punching the air in celebration . 
 Although the band liked the demo , it was difficult for them to record the song . Bassist Adam Clayton said , " At the time it sounded like a foreign language , whereas now we understand how it works " . The arrangement , with two time signature shifts and frequent chord changes , was rehearsed many times , but the group struggled to get a performance they liked . According to co @-@ producer Daniel Lanois , " that was the science project song .
 
 } \\
        \cline{1-2}
         Figure~\ref{fig:graph-2} & {\tiny = Fort Scott National Historic Site = 
 
 Fort Scott National Historic Site is a historical area under the control of the United States National Park Service in Bourbon County , Kansas , United States . Named after General Winfield Scott , who achieved renown during the Mexican @-@ American War , during the middle of the 19th century the fort served as a military base for US Army action in what was the edge of settlement in 1850 . For the next quarter century , it was used as a supply base and to provide security in turbulent areas during the opening of the West to settlement , a period which included Bleeding Kansas and the American Civil War . 
 The current national historic site protects 20 historic structures , a parade ground , and five acres ( 20 @,@ 000 m $^2$ ) of restored <unk> prairie , inside the city of Fort Scott . It is open to visitors most days of the year . 
 
 = = History = = 
 
 In 1842 , Fort Scott was named after Winfield Scott , was established on the American frontier on the military road in eastern Kansas between Fort Leavenworth and Fort Gibson . It was established to provide protection to the rapidly increasing number of settlers , who were migrating from the Eastern United States . Fort Scott became one of a chain of forts intended to protect the new settlers from the Plains Indians , as well as to protect the Indians from the settlers ' encroachment . 
 The United States government intention to reserve permanent Indian lands west of the Missouri River gave way to the competition of settlers continuing to encroach on the Indian settlements . Fort Scott 's most active days were between 1842 and 1853 , although it was also used during the Civil War . 
 
 = = = Army days = = = 
 
 The Cherokee of Indian Territory ( now Oklahoma ) were upset to have Fort Wayne in their proximity . After some delay , the US Army decided to abandon Fort Wayne and move its soldiers to a new fort to be built between Fort Leavenworth and the site . The Army both wanted to placate the Cherokee ( who were supervised by the Department of Defense ) and provide more defense for white settlers and other Indians against the Osage , who had been conducting frequent raids in the area . On April 1 , 1842 , some soldiers of Fort Wayne left their fort and on April 22 arrived where Fort Scott would be built , in the Osage <unk> section of modern @-@ day Kansas . After haggling with the Cherokees to acquire the land , the rest of Fort Wayne 's garrison left the fort on May 26 and arrived at the Fort Scott site on May 30 . 
 Unlike most forts for military use , the fort did not have defensive walls or structures

  } \\

        \cline{1-2}
        \hline
    \end{tabular}
    \caption{Ground truth samples.}
    \label{tab:ground_truth_samples}
\end{table*}

\subsection{Generated examples}
\label{sec:generated_examples}
The generated texts based on the graph shown in \figref{fig:graph-1} and \figref{fig:graph-2} are listed in Table~\ref{tab:samples_vis1} and Table~\ref{tab:samples_vis2}, respectively.

\begin{table*}[h]
    \centering
    \begin{tabular}{p{0.1\linewidth} | p{0.8\linewidth}}
    \hline
        Cond. & Generated Text  \\
        \hline
         - & {\tiny = Final Fantasy Adventure = 
 
 Final Fantasy Adventure is a platform video game developed by Square ( now Square Enix ) and published by Square ( then Square Enix ) for the Super Nintendo Entertainment System , and published by Nintendo for the Nintendo 64 and the Nintendo 64 in North America and Europe . It was released in North America on August 10 , 1995 for the Nintendo 64 and in Europe in March 1995 for the Wii . It was also re @-@ released for the Wii in Japan on March 25 , 2015 for the Wii U 's Virtual Console service . 
 Final Fantasy Adventure is set in the fantasy world of the fictional land of Ooo , between 1000 and 1000 years before the events of the story of the kingdom of Ooo . The story follows the protagonists , Cid , a boy who is sent to Ooo by the legendary Jörmungandr to find the Wizard of Ooo . Throughout the game , players train for a series of random battles , which are played out over a single screen . In the end , players control a " Scarlet Witch " who is summoned to the realm of Ooo . 
 The game was well @-@ received , with praise particularly directed at its graphics and its gameplay , despite criticism for its repetitive gameplay . Critics generally praised its graphics , plot , and simplistic gameplay . The game was commercially successful , selling over 20 @,@ 000 copies , and in 2004 it was re @-@ released in Japan as part of the Final Fantasy Collections series . 
 
 = = Gameplay = = 
 
 The game is a third @-@ person shooter , with gameplay elements related to worlds , time travel , and exploration . Characters are made up of two polygons ; the main character , Aya , is a sprite @-@ based character and inhabits a world of trees and caves , where she is able to perform offensive actions along the way . The protagonist , Aya , is a heavily armored , humanoid creature with red eyes and black eyes . In contrast to other Final Fantasy games , her prominent appearances in this game are not as straightforward as in previous Final Fantasy installments . 
 The game 's world is divided into two sections : the world of Ooo and the world of the dinosaurs . The player 's adventure begins with Aya , a girl who initially leads the plot of Twilight Town , but later returns to Ooo to save her sister . Aya explores the area and is able to save her family . After interacting with the player , Aya is able to resume the story in the future . 
 The game 's main family member , Aya , is a woman named Alice . She is a member of the religious order , the Jehovah 's Witnesses , who is a member of
 
 } \\
        \cline{1-2}
        BoW & {\tiny = Zoo Station ( song ) = 
 
 " Zoo Station " is a song by Australian rock band U2 from their second studio album , PopMart , released in 1992 as the album 's lead single . Rise to popularity after its release prompted U2 to record two songs with bassist Adam Clayton and guitarist Joe Bono , as they felt that it was their breakthrough album . 
 The song comprises lyrics set in an acoustic , with lyrics that have always focused on a woman being stuck in an abusive relationship . It was developed due to the band 's involvement in the construction of Zoo Station and its accompanying soundtrack , and it was inspired by U2 's admiration of the groups they both represented . The song was inspired by a series of incidents that occurred in El Salvador when a man repeatedly attacked an apartment building . The song received positive reviews from music critics , and it was labelled a " masterpiece " by several publications . Thirty @-@ five tracks were released as singles , and the song has been covered by numerous artists , including such figures as Sam \& Max , The <unk> , and Mickey Rourke . The band released their version of " Zoo Station " on non @-@ consecutive studio albums , with " Where the Streets Have No Name " and " Numb " released in 1999 and 2000 respectively . " Zoo Station " is on the band 's greatest hits album , 10 Mile Wild , and on their greatest hits album Across the Universe . It was performed live at the Vertigo Tour in 2007 . The song was included on the band 's fifth studio album , New Order ( 2008 ) , and was included on the film The Last Years ( 2012 ) . 
 
 = = Background = = 
 
 " Zoo Station " was written by U2 frontman Bono and produced by U2 and was the first song on the album . Clayton still had a strong fan base and was a regular singer . The two met while performing in the amphitheaters of Los Angeles , and in a live performance he was the lead singer on the last leg of the Vertigo Tour . The Edge and Clayton were both fans of the band , and the pair decided to collaborate on the album . Both performed on their 2004 tour , and made a solo appearance on the 2004 The Zoo TV Tour . Clayton and Clayton had been close friends , and the pair became friends again in 2008 . In late April 2004 , U2 announced that the song had been released as the first single for the album , and would be released on 31 May , five weeks after the album 's release . " Zoo Station " was released as the fourth single from PopMart and premiered on 13 June 2005 . 
 The song is a
 
  } \\

        \cline{1-2}
        Nodes &  {\tiny= You Can 't Take This Away ( U2 song ) = 
 
 " You Can 't Take This Away ( U2 song ) " is a song by the Irish rock band U2 . It was written by vocalist Bono and produced by The Smiths for their third solo album , The Joshua Tree . Inspired by Romania roots rock and roll , " You Can 't Take This Away " is a song about a man who suffers from severe nightmares . The song was first released on the album and premiered on radio on 19 August 1996 , where it reached number 40 on the Billboard Hot 100 . 
 " You Can 't Take This Away ( U2 song ) " received mixed reviews from critics ; some tracks , such as the opening chorus and bridge , were praised as highlights by some reviewers . The song was well received by critics , as the record company 's highest @-@ selling single at that time , where " The Edge of Forever " peaked at number 64 . It was later re @-@ released on the band 's 2006 compilation album No Line on the Horizon , but has since been re @-@ released on live performances in 2006 and 2009 . " You Can 't Take This Away ( U2 song ) " was performed on the Late Show with David Letterman on 31 December 2005 . 
 " You Can 't Take This Away ( U2 song ) " has since been covered by many groups and has achieved enormous commercial success . A video for the song was filmed by then @-@ frontman Bono , for which it was nominated for a Grammy Award . 
 
 = = Background and writing = = 
 
 " You Can 't Take This Away ( Kingdom of Ireland song ) " is a track that features Bono and The Smiths discussing their relationship and how they changed their lives . His father , Jack Clayton , was assassinated in 1981 at the age of 23 . Bono was born in Philadelphia and worked for a business , first working as a secretary in Los Angeles , then as a photographer for a commercial for Primus . He later worked for the Coca @-@ Cola Company as a drummer in the music industry . The musician picked up the song again after Nikolai <unk> , an engineer who worked with The Smiths , heard it and decided to play it for him after U2 agreed to record it for a solo album . The band originally intended to release " You Can 't Take This Away " as a single ; however , with a critical failure , the song eventually became a single . 
 In 2004 , " You Can 't Take This Away ( U2 song ) " is one of two songs on the album that U2 released as a single with approval of the record label . The
[ ]

} \\
        \cline{1-2}
        GNN & {\tiny = The Edge of Glory = 
 
 " The Edge of Glory " is a song by Irish rock band U2 , released as a single . It was written by guitarist Larry Mullen , Jr . , who plays lead guitar on the song , and was produced by Alex <unk> , who described the song as " a track with a lot of meaning , but no connection . " The song contains several pop rock elements and is set in that time period , and is among the most prominent in the album . In addition to its lyrics , the song 's lyrics detail hypocrisy , and also deals with the effects of adultery . 
 The song 's lyrics have been described by music critics as being autobiographical . The lyrics have been described as " a bold exploration of the figure of a New York City man " , and " an expression of the inability of freedom to live in a world that is also a place in the world of space . " The song 's lyrics describe a " Manhattan @-@ like place " , with Bono calling the arrival a " pleasant little optimism from before it came to life . " 
 " The Edge of Glory " was a success in the United Kingdom , reaching number two in the charts in the United States , and topping the charts in Australia and New Zealand . The song has been certified platinum by the Recording Industry Association of America , and has sold over four million copies worldwide . The song has been covered by several artists , including German band U2 . The music video for " The Edge of Glory " won Best Video at the 2004 MTV Video Music Awards . The video also served as an inspiration for the film U2 360 $^\circ$ ( 1998 ) . 
 
 = = Background = = 
 
 The song has been described as a " relaxed representation " of globalization , with Bono proclaiming himself the " lost king of rock ' n ' roll " , and Chris McGuinness as " the only one who has ever achieved the sound of a rock ' n ' roll . " Bono 's lyrics have been described as a parody of Lord Byron 's " My Own Time " , and as an " attack on social and political issues " . 
 
 = = Recording and production = = 
 
 Bono and U2 made the final stages of recording the song at the Abbey Road Studios in Dublin , Dublin . The sessions were divided into two sessions : Sessions at Damien the flautist and Context at the Cave of Christ . The results of the sessions were mixed by Brian Eno . U2 was very excited to record the result , with Eno referring to the recording as a " special event " , and they decided to change the track

} \\
        \hline
    \end{tabular}
    \caption{Generated samples based on the ``Where the Streets Have No
Name'' graph.}
    \label{tab:samples_vis1}
\end{table*}

\begin{table*}[h]
    \centering
    \begin{tabular}{p{0.1\linewidth} | p{0.8\linewidth}}
    \hline
        Cond. & Generated Text  \\
        \hline
         - & {\tiny = Carbuncle = 
 
 Bloodlines is a 2011 action role @-@ playing video game developed by <unk> and published by Activision for the Wii . The game is part of a sequel of the 2013 game Eclipse : The Series . It was published by Activision on November 19 , 2011 , for Microsoft Windows , Apple II , Xbox 360 , iOS , and Xbox 360 . 
 The game , set in the fictional Middle Ages , revolves around the player @-@ character centaurs battling for control of their universe . The protagonist of the game , Percy , the leader of the Knights of the Old Republic , is a member of the Knights of the Old Republic , and is appointed to lead a military coup to overthrow the Irish Republic and destroy the Home Nations ' military forces . Though the game mainly took place in a new version of the New York City , the original plan was to make it more easily accessible to players unfamiliar with the New Republic . It was also a commercial success , selling more than 900 @,@ 000 copies . 
 The game received mostly positive reviews from most video game publications , with many praising the visual style and the gameplay , but many said that it was not as good as that of the previous game . Reviewers noted the game 's title forward addressing issues such as the difficulty level , a general danger of being too difficult to fight , and the difficulty of playing the game as the player @-@ character 's pattern of character . 
 
 = = Gameplay = = 
 
 Bloodlines is a crossover action role @-@ playing game that takes place in the fictional Middle Ages , which is composed of medieval countries and locales . Valhalla , a medieval stronghold , is the game 's main setting . The player @-@ character is a 3 @-@ D miniature character with a sword and shield , which have multiple colored attacks , and has two of the four abilities , which are progressively reduced from the first one and allow for greater size and movement . The available weapons are bolt @-@ fired weapons , advanced weapons , and weapons that can be used in battle . The player is able to summon magical powers to attack targets , and can use magical powers to enhance the character 's abilities . <unk> are also available via a <unk> system , which enables players to throw stones at enemies and attack enemy characters who have not encountered them . The player character also has an ability to revive foes by performing a touch @-@ screen action . 
 The game can be played as a side @-@ scrolling through a View Mode , which can be used in the single @-@ player mode . The first act features a " <unk> " displayed from a first @-@ person perspective . The player character can move around
 
 } \\
        \cline{1-2}
        BoW & {\tiny = Civil War Pass = 
 
 Civil War Pass , also known as the Battle of the Crater or the Battle of Fort Sumner , was an important battle fought on September 7 , 1864 , at Fort Coldwater , in the state of Montana . 
 After seeing repeated attacks on the fort , Gen. James A. Douglas , the commander of the Confederate forces in the South , decided to abandon the fort and flee to the north . After Union forces struck the fort , they decided to flee south to the Ohio River . There they quickly encountered a group of horses , who were used to build a pontoon bridge . The ditches and wooden planks were removed and replaced with stone blocks to make them float ( plow ) . The obstacles that were created in the river valley , however , proved treacherous and were not bridged by mountain passes . The young general and his troops eventually reached the Ohio and the Mississippi rivers , but the new Presidential candidate , Abraham Lincoln , resigned after the war . 
 After the defeat at Fort Sumner , General Douglas , the commander of the Union forces , planned and executed a number of attacks on Fort Sumner . When soldiers arrived , they found two now @-@ deserted locations . The attacks had been made more than a year before . When the line of retreat of the Union forces , which stretched from Fort Sumner to Fort Sumner , reached Fort Sumner on August 19 , 1864 , the cavalrymen captured it on September 30 . In November 1864 , General Douglas was defeated at the Battle of Lake Logan .
 
 = = Background = = 
 
 In 1861 , with the Mexican @-@ American War nearing its conclusion , the American public began to think of an armistice treaty , or peace treaty between Mexico and the United States . On July 1 , 1861 , General Douglas sent three large armies from the Mexican @-@ American War , a series of forts west of the Rockies , to attack Fort Vicksburg . The forts were destroyed in a siege in June . These were built during the years it was fought by the Confederate States of America . The British and Americans were unprepared for the chance of victory , and the Americans were now planning to take control of the Gulf Coast . Like the Americans , the British were planning an attack into central Canada . The British were aware that the main invasion of Canada would occur on July 8 . The British were near the Niagara River and the Union were hopefully midway along the river , approaching Fort Sumner from the west . The British were reluctant to move toward the Carolinas , and so , in the event the Port of Boston was abandoned , the British would be forced to travel to the lower Mississippi . 
 The
 
  } \\

        \cline{1-2}
        Nodes &  {\tiny = Fort Scott = 
 
 Fort Scott is an American military post located in Fort Lee , Kansas . It is named in honor of General William Scott , a U.S. Army general and the first commander of the Army of the Potomac . The site was designated as a National Historic Landmark in 1991 , and has been designated a National Historic Landmark under the title of Fort Scott Historical Site since 1929 . It is located in the Rocky Mountains in Kansas and is known as the " James Scott National Historic Site " . 
 
 = = History = = 
 
 The original having been settled by the Caddo on the Black River , and later moved to Fort Lee in present @-@ day Decatur County , Virginia . On July 10 , 1810 , the Hennepin reported that the Caddo had acquired the territory of Fort Lee , but it is unclear whether he was present there . He may have taken a position that had previously been occupied by other people . Around 1800 , the first Governor of Kansas , Colonel Andrew H. Sharpe , established Fort Scott in what is now a part of Fort Lee . The fort was constructed on a site that he had named Fort Scott , and was known as Fort Douglas . The fort was used for administrative purposes and for administration of the Missouri Territory . 
 In 1808 , William Bolivar Buckner led a large movement to remove the western boundary of Texas , including Fort Scott . Congress authorized a survey of the territory in 1817 , and a survey of the Old South boundary was completed in 1818 , making Fort Scott the first governor to apply federal law . Although the West Texas Aftermath quickly became a national concern , the new governor was unable to raise sufficient funds to maintain Fort Scott . 
 The fort 's construction and construction were completed in 1821 , but the state legislature refused to grant the commissioners the land they were granted . The new land , called Fort Dix , was consequently purchased by the U.S. Army . The fort 's name was later changed to Fort Lee . 
 While the two states were in dispute by the 1832 treaty , Fort Dix was located in the Horn of the Midget Valley , part of the Pan @-@ American Native Reservation . Confederate forces launched a cavalry attack on Fort Dix in early 1835 , but both sides suffered defeats to the Union . Fort Dix was declared a U.S. Army national monument by President Andrew H. Sharpe in September 1836 . 
 Fort Dix was named after General John Scott , a U.S. Army general and the first governor of Texas , who was killed in an assassination attempt on June 20 , 1855 . Military historian John P. Pickett wrote that it was the first military governor in the United States to serve in the Confederate States
 
} \\
        \cline{1-2}
        GNN & {\tiny = Fort Scott National Historical Park =
        
 Fort Scott National Historical Park is an Illinois state historic park in the U.S. state of Kentucky . It is located at $53 ^\circ 25 ^\prime 4 ^{\prime\prime}$ N $65 ^\circ 41 ^\prime 16 ^{\prime\prime}$ W , at the edge of the Clay Creek valley , southwest of New Orleans . It is located at the intersection of Washington Boulevard and State Route 63 , and is the largest National Historic Landmark in the state . The site was purchased by Native Americans in 1803 and the site was added to the National Register of Historic Places in 1962 . Since 1998 , the site has been subject to an extensive series of historic markers and features that are important in preservation of American historic sites in Texas . 
 The National Park Service includes the nation 's oldest extant log cabins , historic buildings , historic facilities , and historic structures . The park is home to the Mississippi River National Historical Park , a U.S. National Monument that supplies historic sites and historic sites . The original fort was built in 1818 to protect U.S. statehood . In 1899 , the state legislature constructed a small blockhouse at the site of the original fort to defend it from Native Americans . The blockhouse first appeared in 1868 , when land in the city of Lisbon was granted to the state . The fort has remained in use since then .
 
 = = History = = 
 
 = = = Early history = = = 
 
 Fort Scott was established as a civil and military fortification in 1803 and named after an American Indian . The land that would become Fort Scott was originally part of the Louisiana Purchase , which was granted to the United States by the Louisiana Purchase Act of 1825 . The original fort was established in 1828 by an act of Congress . The American Revolutionary War came to an end in 1830 , but Independence was declared in 1831 and Independence was declared on June 3 , 1830 . The post @-@ war Treaty of Paris signed at Fort Scott ended military activity in the region . War by the United States reached an end in 1830 , and most of the land was put aside for use as a military park . 
 Fort Scott was garrisoned by 90 soldiers from the 55th Louisiana Regiment during the War of 1812 . In 1837 , the Illinois General Assembly passed legislation creating Fort Scott as a federal park , and in the same year the state agreed to purchase the site in honor of the site 's new state of Louisiana . Originally , only about half of Fort Scott was owned , but the size of the park changed in the 1880s from a forest reserve to a dirt road . The park was significantly expanded during the 1910s , but the exact date is disputed . The
 
} \\
        \hline
    \end{tabular}
    \caption{Generated samples based on the ``Fort Scott National Historic Site'' graph.}
    \label{tab:samples_vis2}
\end{table*}

\subsection{Ablations on sampling configurations}
\label{sec:appendix_sampling}
We show additional ablation results on the sample length (Table~\ref{tab:sampling_length}) and the temperature  (Table~\ref{tab:sampling_temperature}) for greedy sampling. Note that for each case we show the rBLEU score based on the validation set computed with a single sampling run (20 samples per graph).

Note that the GNN model has overall the best performance.  However as the sample length increases the advantage of the GNN model also decreases.  This indicates that it is still very challenging to generate long text that stays on-topic, and potentially the noise overwhelms the signal when number of tokens increases to 4096.

\begin{table*}[h]
    \centering
    \begin{tabular}{c|c|c|c|c|c|c|c|c|c|c}
    \hline
        \multirow{3}{*}{Cond.} & \multicolumn{5}{c|}{Valid rBLEU} & \multicolumn{5}{c}{Valid rBLEU (w/ title)} \\ \cline{2-11}
        & \multicolumn{5}{c|}{Sample length} & \multicolumn{5}{c}{Sample length} \\ \cline{2-11}
         & 256 & 512 & 1024 & 2048 & 4096 & 256 & 512 & 1024 & 2048 & 4096 \\
        \hline
        None & 9.53 & 10.47 & 12.22 & 14.57 & 14.60 & 29.03 & 27.78 & 27.02 & 27.24 & 26.94  \\
        BoW & 30.63 & 29.44 & 29.56 & 29.92 & 30.00 & 35.03 & 32.48 & 31.50 & 31.72 & 31.46 \\
        Nodes & 32.33 & 30.30 & 29.82 & 30.43 & 29.91 & 35.45 & 32.88 & 31.57 & 31.79 & 31.03 \\
        GNN & 33.81 & 31.32 & 30.39 & 30.53 & 30.05 & 36.49 & 32.49 & 31.70 & 31.77 & 30.79 \\
    \hline
    \end{tabular}
    \caption{Generated samples vs sample length.
    }
    \label{tab:sampling_length}
\end{table*}

\begin{table*}[h]
    \centering
    \begin{tabular}{c|c|c|c|c|c|c}
    \hline
        \multirow{3}{*}{Cond.} & \multicolumn{3}{c|}{Valid rBLEU} & \multicolumn{3}{c}{Valid rBLEU (w/ title)} \\ \cline{2-7}
        & \multicolumn{3}{c|}{Temperature} & \multicolumn{3}{c}{Temperature} \\ \cline{2-7}
         & 0.6 & 0.8 & 1.0 & 0.6 & 0.8 & 1.0 \\
        \hline
        None & 12.08 & 10.47 & 9.71 & 27.09 & 27.78 & 26.21 \\
        BoW & 28.21 & 29.44 & 27.63 & 31.25 & 32.48 & 31.02 \\
        Nodes & 29.55 & 30.30 & 28.48 & 31.52 & 32.88 & 31.23 \\
        GNN & 29.59 & 31.32 & 29.01 & 31.55 & 32.49 & 31.20 \\
    \hline
    \end{tabular}
    \caption{Generated samples vs temperature.}
    \label{tab:sampling_temperature}
\end{table*}



\end{document}